# Cross-variable Linear Integrated ENhanced Transformer for Photovoltaic power forecasting

Jiaxin Gao [1,2], Qinglong Cao [1,2], Yuntian Chen [1,2 *], Dongxiao Zhang [2,3,4]

1 Shanghai Jiao Tong University, Shanghai, P. R. China;
2 Ningbo Institute of Digital Twin, Eastern Institute of Technology, Ningbo, Zhejiang, P. R. China;
3 Department of Mathematics and Theories, Peng Cheng Laboratory, Guangdong, P. R. China;
4 National Center for Applied Mathematics Shenzhen (NCAMS), Southern University of Science and Technology, Guangdong, P. R. China.

**ABSTRACT**

Photovoltaic (PV) power forecasting plays a crucial role in optimizing the operation and planning of PV systems, thereby enabling efficient energy management and grid integration. However, un certainties caused by fluctuating weather conditions and complex interactions between different variables pose significant challenges to accurate PV power forecasting. In this study, we propose PV-Client (Cross-variable Linear Integrated ENhanced Transformer for Photovoltaic power forecasting) to address these challenges and enhance PV power forecasting accuracy. PV-Client employs an ENhanced Transformer module to capture complex interactions of various features in PV systems, and utilizes a linear module to learn trend information in PV power. Diverging from conventional time series-based Transformer models that use cross-time Attention to learn dependencies between different time steps, the Enhanced Transformer module integrates cross-variable Attention to capture dependencies between PV power and weather factors. Furthermore, PV-Client streamlines the embedding and position encoding layers by replacing the Decoder module with a projection layer. Experimental results on three real-world PV power datasets affirm PV-Client's state-of-the-art (SOTA) performance in PV power forecasting. Specifically, PV-Client surpasses the second-best model GRU by 5.3% in MSE metrics and 0.9% in accuracy metrics at the Jingang Station. Similarly, PV-Client outperforms the second-best model SVR by 10.1% in MSE metrics and 0.2% in accuracy metrics at the Xinqingnian Station, and PV-Client exhibits superior performance compared to the second-best model SVR with enhancements of 3.4% in MSE metrics and 0.9% in accuracy metrics at the Hongxing Station.

**NONMENCLATURE**

| Abbreviations | |
|---|---|
| PV | Photovoltaic |
| PV-Client | Cross-variable Linear Integrated ENhanced Transformer for Photovoltaic power forecasting |

## 1. INTRODUCTION

Photovoltaic (PV) power, as a clean and renewable energy source, has gained significant attention in recent years driven by its potential to curtail carbon emissions and diminish the reliance on traditional fossil fuels [1]. The efficient utilization of PV energy relies heavily on accurate forecasting of PV system output. Accurate PV power forecasting enables effective power grid planning, load balancing, and resource management, contributing to the overall stability and efficiency of energy systems. Additionally, PV power forecasting facilitates the integration of PV energy into the existing power grid infrastructure, enabling the optimal utilization of renewable energy sources while ensuring grid reliability and stability [2,3]. As we navigate towards a future dominated by sustainable energy, the role of accurate PV power forecasting stands as a pivotal element in achieving a harmonious coexistence between renewable sources and the established energy infrastructure.

---

* Corresponding Author.

Numerous research studies have been conducted to devise accurate and computationally efficient forecasting models for PV power generation. These models can be broadly categorized as indirect and direct forecasting models. Within the realm of indirect forecasting models, various methods including numerical weather prediction (NWP) [4,5], statistical approaches [6], and image-based methods [7] have been utilized to predict solar radiation at different time scales. Subsequently, these forecasted solar radiation values, along with other relevant data, serve as inputs to estimate the PV power generation. In contrast, the direct forecasting model directly predicts the PV power generation using historical PV power and associated meteorological data on the basis of relatively accurate weather forecast data. The spectrum of direct forecasting models encompasses persistence models [8,9], statistical models [10], machine learning models [11–13], and hybrid models [14–16]. The selection between indirect and direct forecasting models depends on factors such as data availability, computational resources, and specific forecasting requirements. As the demand for refined PV power forecasting methodologies continues to grow, these diverse forecasting models serve as adaptable instruments, accommodating a spectrum of requirements within the realm of renewable energy prediction.

Despite notable progress in the field of PV forecasting, several challenges persist, making accurate predictions a complex undertaking. A significant challenge stems from the inherent variability and uncertainty in weather conditions, given that PV system output is heavily contingent on solar radiation levels. The rapid fluctuations in solar radiation and daily patterns pose difficulties in accurately capturing and modeling short-term and long-term variations [17]. Additionally, other weather factors such as temperature, cloud cover, surface pressure, and other atmospheric conditions can also have impacts on the production of PV power [18]. The inherent inaccuracy of weather forecasts further compounds this challenge. Another challenge lies in the non-linear relationships between weather factors and PV power output [19]. Conventional linear models often fall short in capturing the intricate interactions and dynamics present in PV systems, resulting in less precise predictions. Addressing these challenges and developing accurate PV power forecasting methods is crucial for ensuring the dependable integration and utilization of PV energy. However, previous models for PV series forecasting have primarily concentrated on the interdependencies between different time steps within the series, which results in an excessively local receptive field, neglecting the holistic attributes of PV series or weather variables. Consequently, the model fails to fully capture the trend information of PV series and their dependencies on weather variables.

In addressing these challenges, we propose PV-Client (Cross-variable Linear Integrated ENhanced Transformer for Photovoltaic power forecasting), which is an adaptation of the time series model Client [20], specifically tailored for PV power forecasting. PV-Client learns from both historical PV power data and weather forecast data, providing accurate predictions of PV power. PV-Client is positioned as a direct forecasting model, and this choice is driven by the availability of a reasonably extensive historical dataset and the relatively high accuracy of weather forecast information. PV-Client adeptly captures intricate interactions within PV systems through the utilization of an Enhanced Transformer module. In this module, distinct variables (such as PV power or weather factors) are treated as separate entities for modeling. Utilizing an efficient Attention mechanism [21], the dependencies among these variables can be more effectively captured. This stands in contrast to traditional time series-based Transformer models, where the Attention mechanism is typically employed to capture dependencies between different time steps [22,23]. Additionally, PV-Client integrates a linear module to discern trend information in PV power. The synergistic integration of linear and non-linear components equips PV-Client to adeptly capture both global and local patterns in PV power: Cross-variable attention, functioning as a global mechanism, considers information from all variables, while the linear module focuses solely on the corresponding variable itself, which can be considered local from a feature perspective. This integration elevates PV-Client's efficacy as an efficient tool for precise PV power forecasting. Furthermore, PV-Client incorporates a reversible instance normalization (RevIN) [24] module to mitigate distribution shift [25] in PV power, thereby enhancing the stability of the model's predictions. By comprehensively understanding the nuanced relationships within the PV system, PV-Client enhances its predictive capabilities, addressing the inherent challenges associated with accurately modeling the dynamic nature of PV power generation. The effectiveness of PV-Client is substantiated through real-world experiments conducted at the Photovoltaic Power stations of Jingang, Xinqingnian, and Hongxing.

In summary, our contribution is three-fold:
- PV-Client innovates by improving the structure of time series-based Transformers. It introduces cross-variable Attention in lieu of traditional cross-time Attention, providing a more versatile and robust mechanism for capturing dependencies within the PV systems.
- PV-Client incorporates a linear module to learn trend information in PV power and utilizes an Enhanced Transformer module to capture the intricate interactions within PV systems.
- Through experiments conducted on three real-world PV power datasets, PV-Client has demonstrated state-of-the-art (SOTA) performance in PV power forecasting, showcasing a 5.3% enhancement in MSE metrics and a 0.9% improvement in accuracy metrics compared with the second-best model GRU in Jingang Station, a 10.1% enhancement in MSE metrics and a 0.2% improvement in accuracy metrics compared with SVR in Xinqingnian Station, and a 3.4% enhancement in MSE metrics and a 0.9% improvement in accuracy metrics compared with SVR in Hongxing Station.

## 2. METHODOLOGY

**Preliminary.** Given a historical PV series **S={$x_1,...,x_L$}** and weather forecast data **W={$w_T,...,w_{L+T}$}**, where *L* denotes the length of the historical PV series and *T* represents the number of time steps to forecast, the objective of PV power forecasting is to predict future PV power production **F** based on **S** and **W**, denoted as **F={$x_{L+1},...,x_{L+T}$}**.

The core concept underpinning PV-Client involves the incorporation of cross-variable Attention in the Enhanced Transformer module to learn the dependencies between PV power and weather factors, departing from the cross-time Attention employed in conventional time series-based Transformer models. Additionally, the model seamlessly integrates a linear module to capture the inherent trends in PV power. Furthermore, PV-Client employs a RevIN [24] module to mitigate the PV series' distribution shift. The overall architecture of the PV-Client can be elucidated by referring to Fig. 1. These architectural enhancements are intended to facilitate more effective utilization of variable dependencies and trend information in PV power. This section provides a detailed exposition of the individual components constituting the PV-Client model.

### 2.1 Cross-variable Transformer

The Attention mechanism in the cross-variable Transformer module is employed to capture variable dependencies, specifically between PV power and weather factors, in place of time dependencies in PV power, as shown in Fig. 2. From this perspective, each time step is no longer viewed as a cohesive entity; instead, each variable (PV power or weather factors) is treated as an individual entity, engaging in interactions with other individual entities.

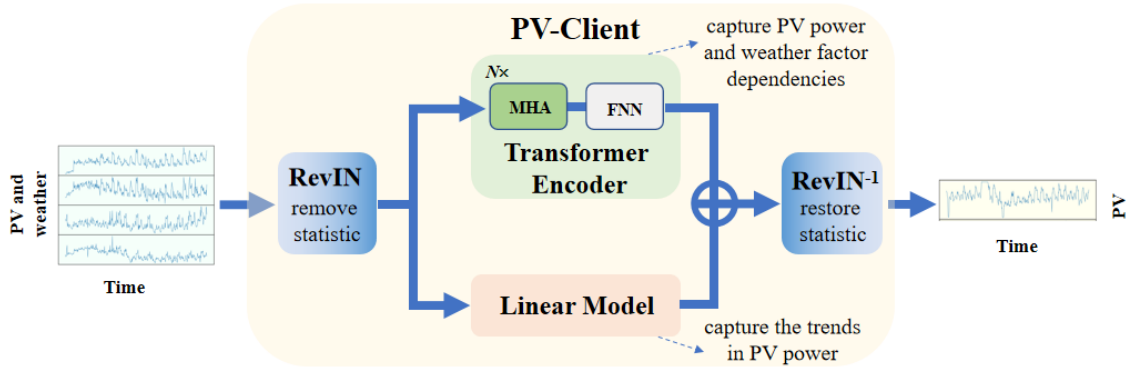

Fig. 1 PV-Client's architecture. The RevIN module is used to address the issue of distribution shift of PV power series. The linear model is used to capture trend information, while the enhanced Transformer model is used to capture nonlinear and cross-variable dependencies in PV power series.

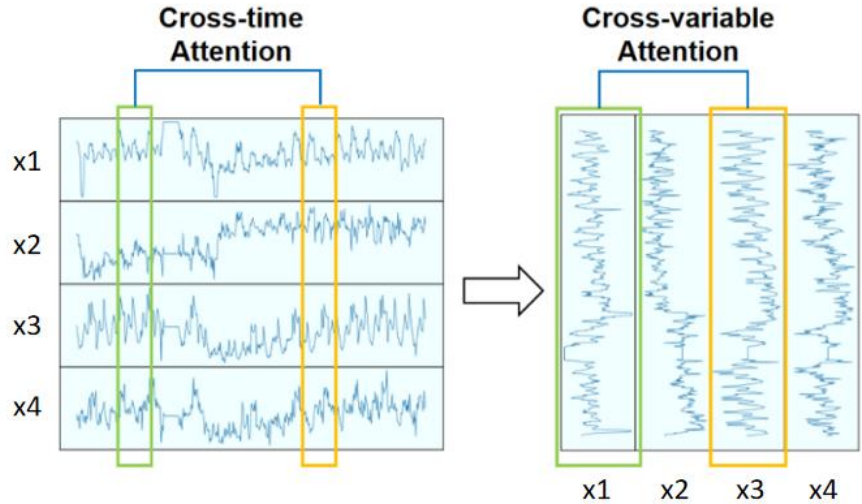

Fig. 2 Cross-variable Attention, and x1-x4 represent different channels in multivariate time series. The Attention module is employed to capture dependencies between PV power and weather factors, rather than dependencies among different time steps within the PV power series.

The Transformer's Encoder block comprises a multi-head Attention (MHA) component and a feed-forward network (FFN) component, as shown in Fig.1. The input series **H**, composed of the historical PV power **S** along with the weather forecast data **W**, is a 2D Tensor with the shape of $L \times C$, where $L$ represents the time steps of the input, and $C$ denotes the number of variables. The cross-variable Attention is the key part of MHA, which can be defined as Eq. (1):

$$\text{Attention}(\mathbf{Q}, \mathbf{K}, \mathbf{V}) = \text{softmax}\left(\frac{\mathbf{Q}\mathbf{K}^\top}{\sqrt{L}}\right)\mathbf{V} \tag{1}$$

where **Q** is queries, **K** is keys, and **V** is values. **Q**, **K** and **V** are generally obtained by applying some transformations to the input of MHA, and *L* is length of the input. By assigning distinct Attention weights to different variables within the PV power series data and weather factors, the Attention module is widely acknowledged as an effective tool for capturing the intricate dependencies among these variables. Unlike the conventional linear or MLP layers, which are limited in their ability to model complex relationships between variables, the Attention mechanism allows for dynamic and flexible modeling of such dependencies.

MHA divides the conventional Attention module into multiple heads, enabling the module to capture diverse patterns and relationships more effectively. By splitting the input into multiple heads and computing Attention independently for each head, MHA allows the module to attend to different parts of the input simultaneously, thus facilitating the extraction of intricate features and dependencies. This capability enhances the effectiveness of the model with MHA components across various tasks [26,27].

The Feed-Forward Network (FFN) module in Fig.1 serves as a crucial component for nonlinear transformation and feature extraction within the Enhanced Transformer's Encoder block. Consisting of two fully connected layers followed by an activation function, typically ReLU [28], the FFN module is designed to map feature vectors at each position to a higher-dimensional space before projecting them back to the original space. This process enhances the representational capacity of features, enabling the model to capture intricate patterns and relationships within the PV power series and weather factors more effectively. Such nonlinear transformations significantly contribute to the improved performance of the Enhanced Transformer model.

The input series is directly fed into the Transformer Encoder blocks after being flipped, bypassing the use of an embedding layer typically present in the Transformer model, as shown in Fig.1. In the conventional Transformer Encoder blocks, the embedding layer is usually employed to map discrete input tokens into continuous vector representations, facilitating the model's learning of meaningful relationships between different tokens in the input [29]. However, for time-series data, the input is not composed of discrete tokens. Instead, each time step is strongly correlated with other time steps, rendering the embedding layer unnecessary. Its removal prevents compromising temporal information and avoids potential performance degradation. Additionally, the position encoding layer [30] in the conventional Transformer is omitted, considering the absence of temporal ordering among different variables. Specifically, in PV power forecasting, various weather factors represent different variables, and rearranging the order of these variables does not affect the forecasting results.

Following feature extraction from the Encoder blocks, the processed series is directed to a projection layer. The output, representing the prediction of the cross-variable Transformer, is obtained without involving a Decoder block. This is motivated by the observation that including a Decoder block results in reduced performance. From our perspective, the cross-variable Transformer module in the PV-Client model primarily functions as a feature extractor rather than a series generator, and it inherently does not encode temporal relationships among different variables. Therefore, we deem a Decoder unnecessary in this context. The process of projection is defined as:

$$\mathbf{F}_{\text{trans}} = \text{Proj}(\mathbf{X}_{\text{enc}}).\text{Permute}(1,0) \tag{2}$$

where $\mathbf{X}_{\text{enc}}$ signifies the output of Encoder blocks, and $\mathbf{F}_{\text{trans}}$ represents the Transformer's prediction. The notation 'Permute(1,0)' denotes the permutation of dimensions 1 and 0, which indicates the flipping of the output of the projection layer along these dimensions. After the flipping operation, the shape of the output aligns with the shape of the input. The cross-variable Transformer's prediction often contains the details of the PV power.

*2.2 Linear Integration and RevIN Modules*

The integrated linear module is employed to extract trend information from the PV power series by leveraging its capability to capture linear relationships and gradual changes inherent in the PV power series over time. The integrated linear module operates in a channel-independent manner [31], where the prediction of PV power depends solely on the historical series of PV power and remains unaffected by other weather variables. Notably, the impact of weather variables on PV power has been sufficiently captured by the Attention module within the cross-variable Transformer block. By leveraging these insights, the proposed framework minimizes the risk of noise introduced by redundant learning and avoids redundant processing of the linear module. The input series is flipped and smoothed by the RevIN module, and put into the linear module to get the linear module's prediction, as defined:

$$\mathbf{F}_{\text{lin}} = \text{Linear}\big(\mathbf{O}.\text{Permute}(1,0)\big).\text{Permute}(1,0) \qquad (3)$$

where $\mathbf{O}$ represents the smoothed output of the input $\mathbf{H}$ processed by the RevIN module, and the notation 'Permute(1,0)' also signifies the permutation of dimensions 1 and 0. The cross-variable Transformer's prediction and the linear module's prediction are combined with learnable weights $\mathbf{w}_{\text{trans}}$ and $\mathbf{w}_{\text{lin}}$ to get the combined prediction of PV power, as described in:

$$\mathbf{F} = \mathbf{w}_{\text{trans}} \times \mathbf{F}_{\text{trans}} + \mathbf{w}_{\text{lin}} \times \mathbf{F}_{\text{lin}} \qquad (4)$$

where $\mathbf{w}_{\text{trans}}$ and $\mathbf{w}_{\text{lin}}$ are updated along with the other parameters of the model, allowing for the dynamic learning of the relative weights between the trend and details within the PV series.

Due to the instability of weather factors such as radiation, PV series data often encounter a distribution shift problem, characterized by changes in statistical properties like mean and variance over time. This challenge significantly impacts the accuracy of PV power forecasting. To address the issue of distribution shift in PV power, a reversible instance normalization (RevIN) [24] module is adopted in the model, which is symmetrically structured to remove and restore the statistical information of a PV power series instance. The process of instance normalization can be defined as:

$$\mathbf{O} = \text{RevIN}(\mathbf{H}) = \boldsymbol{\alpha} * ((\mathbf{H} - \boldsymbol{\mu}) / \boldsymbol{\sigma} + \boldsymbol{\beta} \qquad (5)$$

where $\mathbf{H}$ is the input of PV-Client, $\boldsymbol{\mu}$ denotes the mean value of the input instance, and $\boldsymbol{\sigma}$ represents the standard deviation of the input instance. The parameters $\boldsymbol{\alpha}$ and $\boldsymbol{\beta}$ are learnable affine parameter vector. After the instance normalization process with the RevIN module, the smoothed series is directed to both the cross-variable Transformer module and the linear module. Subsequently, the outputs of these two modules are combined with weighted sums. The combined prediction undergoes denormalization via the RevIN module to get the final prediction. The process of instance denormalization can be described as follows:

$$\mathbf{F}' = \text{RevIN}^{-1}(\mathbf{F}) = (\mathbf{F} - \boldsymbol{\beta}) / \boldsymbol{\alpha} * \boldsymbol{\sigma} + \boldsymbol{\mu} \qquad (6)$$

where **F** is the combined prediction of the Enhanced Transformer module and the integrated linear module, and **F′** represents the final prediction of PV-Client. The instance normalization and denormalization process performed by the RevIN module can promote the model's stability during forecasting.

*2.3 Overall PV-Client Architecture*

PV-Client integrates an enhanced Transformer module to capture nonlinear information and cross-variable dependencies within PV power, and a linear module to capture trend information. This conceptual framework shares similarities with some theory-guided time-series forecasting models [32,33]. The distinction lies in the fact that theory-guided models acquire temporal trend through domain knowledge, whereas PV-Client autonomously learns trend information through its linear module. As illustrated in Fig.1, the input series is first smoothed with the RevIN module. Then, the smoothed series is fed into both the cross-variable Transformer module and the linear module. The final PV power prediction combines outputs from both modules and is denormalized using the RevIN module. This approach ensures that PV-Client captures both global and local patterns in PV series data, enhancing its accuracy and reliability in real-world applications.

The original Client model is designed to forecast multivariate time series, hence its output consists of multiple dimensions [20]. However, for the task of PV power forecasting, the primary focus lies in predicting PV power exclusively, without the necessity of forecasting the other weather variables. Consequently, a single variable dimension suffices for the PV-Client's output to denote the PV power prediction. There are three options to connect the variable dimension of its output to the input dimensions: The first option involves connecting the variable dimension of the output directly to the PV power dimension of the input, which is also the most straightforward approach. Secondly, linking the variable dimension of the model's output to the most relevant weather variable, specifically the radiation forecast dimension of the input, presents an alternative option. Thirdly, computing a weighted sum of the outputs from the PV power dimension and the radiation forecast dimension provides an additional choice. Through experimentation in subsection 3.4, it has been determined that the first option yields the most effective results, which is logical. In the PV-Client's cross-variable Transformer module, information across different input dimensions is shared. However, for the integrated linear module within the PV-Client, trend information learned solely from the PV power series itself proves to be more accurate. Despite the high correlation between radiation forecast and PV power, radiation forecast inherently contain biases, and the trend of radiation forecast cannot precisely mirror the trend of PV power.

### 3. EXPERIMENT

*3.1 Data description and experiment setting*

In this study, we analyze the PV power data from three real-world stations: Jingang Photovoltaic Power Station in ShenZhen, China; Xinqingnian Photovoltaic Power Station in Ningbo, China; and Hongxing Photovoltaic Power Station in Sanya, China. These three power stations are located in southern China, where weather conditions fluctuate frequently, posing significant challenges to PV power forecasting. In the experiments, PV-Client utilizes historical PV power data from the preceding two days and weather forecast data, including radiation, temperature, humidity, wind speed, and surface pressure. While radiation strongly influences PV power output, it's worth noting that other weather factors may also have an influence on it. Compared to the authentic weather data obtained from meteorological monitors, weather forecast data exhibits similar trends but may differ, especially in capturing short-term fluctuations. Fig. 3 illustrates the comparison between the radiation forecast and the actual radiation. It is evident that significant disparities exist between radiation forecasts and actual radiation in terms of magnitude and volatility.

As both the PV power and weather forecast data are sampled every 15 minutes, the input length is 192 with a feature dimension of six. The objective is to predict the PV power for the following day, as day-ahead forecasts play a crucial role in power generation scheduling. The training data is standardized, and during testing, the testing data is standardized using the normalization parameters obtained from the training data. After obtaining the model's predictions, the same parameters used for normalization are applied to inversely standardize the model's predictions. Concerning model training, default values commonly used in machine learning are adopted as the values for hyperparameters. Specifically, the number of the Enhanced Transformer Encoder layers is configured to 2, and the hidden state dimension is set to 128. The ADAM optimizer [34] with a learning rate of 1e-3 is employed. The batch size is specified as 128, and the training epoch is set to 10. The initial weight for the Transformer model $\mathbf{w}_{trans}$ and the initial weight for the linear model $\mathbf{w}_{lin}$ are both set to 1.

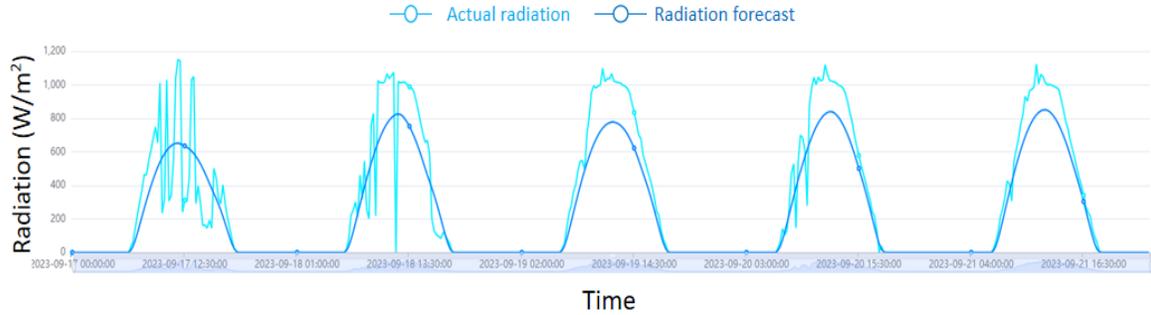

Fig. 3 Comparison of the radiation forecast and the actual radiation in Jingang Station.

*3.2 PV power forecasting results*

For the PV power forecasting experiments, the models are trained using nearly a year of historical offline data, spanning different periods for each photovoltaic power station: from 2022.8.15 to 2023.7.28 for the Jingang Photovoltaic Power Station; from 2022.8.15 to 2023.10.8 for the Xinqingnian Photovoltaic Power Station; and from 2023.3.1 to 2023.10.8 for the Hongxing Photovoltaic Power station. Subsequently, online testing and continuous monitoring of results are conducted for one month, specifically from 2023.9.10 to 2023.10.10 for the Jingang Station, and from 2023.10.10 to 2023.11.10 for the Xinqingnian and Hongxing Stations. The PV power and weather factors are sampled every 15 minutes, resulting in 33,312, 40,224, and 14,304 training samples for Jingang, Xinqingnian, and Hongxing Stations, respectively. For testing, there are 2,880, 2,976, and 2,976 samples for Jingang, Xinqingnian, and Hongxing Stations, respectively. This online evaluation approach aims to offer a more objective assessment of the models' performance in real-world scenarios.

To evaluate the models, we utilize both the mean square error (MSE) and accuracy metrics. It is noteworthy that predictions and actual values are used in their original scale rather than normalized data for the evaluation. The accuracy metric is used to assess the relative error of each model's predictions, as defined in:

$$\text{Acc} = 1 - \frac{\sqrt{\sum_{i=1}^{n}(G_i - P_i)^2}}{\text{Cap}\sqrt{n}} \tag{7}$$

where $G_i$ represents the actual PV power output for a given time step i, while $P_i$ represents the predicted PV power output, and 'Cap' represents installed capacity of PV power plants.

Typically, the changes in the MSE and accuracy metrics are aligned. When optimizing the forecasting models, the focus tends to be on the MSE metric. However, in practical deployment or when reporting to users, more emphasis is often placed on the accuracy metric due to its greater intuitiveness.

We compare six baseline models: Linear Regression (LR) [35], Support Vector Regression (SVR) [36], XGBoost [37], LightGBM [38], Gated Recurrent Unit (GRU) [39], as well as a cross-time based Transformer model (T-Transformer for short). LR is a fundamental and extensively utilized statistical model designed to establish a linear relationship between input variables and the output variable. SVR is a regression model that seeks to identify a hyperplane within a high-dimensional feature space, effectively separating data points and minimizing regression line error. XGBoost and LightGBM are both prominent gradient boosting [40] frameworks that employ an ensemble learning technique. They create a strong predictive model by sequentially adding weak models and focusing on the data points with higher residuals. However, they differ in their implementation details and performance characteristics. XGBoost is implemented in C++ and focuses on pre-sorting and cache-aware learning, while LightGBM utilizes histogram-based algorithms and gradient-based one-sided sampling to enhance training speed and reduce memory usage. GRU is a type of recurrent neural network that utilizes gating mechanisms to selectively update and forget information in the hidden state, allowing it to capture long-term dependencies and make accurate predictions for time series data. The model T-Transformer utilizes an efficient Attention mechanism to capture temporal dynamics and dependencies across different time steps in time series data.

Table 1. MSE and accuracy of different models in Jingang Station. The best results are indicated in bold font.

| Model | MSE metrics | Accuracy metrics |
| --- | --- | --- |
| LR | 1693.707 | 0.894 |
| SVR | 2043.094 | 0.887 |
| XGBoost | 1879.130 | 0.890 |
| LightGBM | 1952.271 | 0.889 |
| GRU | 1555.194 | 0.881 |
| T-Transformer | 4167.539 | 0.835 |
| **PV-Client (ours)** | **1469.132** | **0.903** |

Table 2. MSE and accuracy of different models in Xinqingnian Station.

| Model | MSE metrics | Accuracy metrics |
| --- | --- | --- |
| LR | 20.720 | 0.944 |
| SVR | 7.118 | 0.967 |
| LightGBM | 9.299 | 0.962 |
| **PV-Client (ours)** | **6.396** | **0.969** |

The prediction MSE and accuracy metrics for various models across the three stations are detailed in Table 1, Table 2, and Table 3, respectively. PV-Client notably outperforms other models, showcasing the most favorable results in both MSE and accuracy metrics across all three stations. Specifically, in Jingang PV power Station, PV-Client achieves a 5.3% improvement in MSE metrics and a 0.9% improvement in accuracy metrics compared to the second-best model. In Xinqingnian PV power Station, the improvements are 10.1% in MSE metrics and 0.2% in accuracy metrics, while in Hongxing PV power Station, PV-Client shows improvements of 3.4% in MSE metrics and 1.6% in accuracy metrics compared to the second-best model. On average, PV-Client exhibits a 6.3% improvement in MSE metrics and a 0.9% improvement in accuracy metrics across the three stations.

Table 3. MSE and accuracy of different models in Hongxing Station.

| Model | MSE metrics | Accuracy metrics |
|---|---|---|
| LR | 76.994 | 0.831 |
| SVR | 75.331 | 0.834 |
| LightGBM | 77.473 | 0.829 |
| **PV-Client (ours)** | **72.741** | **0.850** |

Figures 4-6 visually demonstrate the exceptional prediction performance of PV-Client in Jingang Station, Xinqingnian Station, and Hongxing Station. They also provide comparisons of the predictions made by LR and SVR. The horizontal axis in the figures represent time, while the vertical axis represent the values of PV power. The red line corresponds to the actual PV power values, the light blue line represents the predictions made by PV-Client, the dark blue line denotes the LR's predictions, and the purple line represents the SVR's predictions. As shown in the figures, PV-Client's predictions are much closer to the actual PV power output in most cases when compared to the predictions made by LR and SVR.

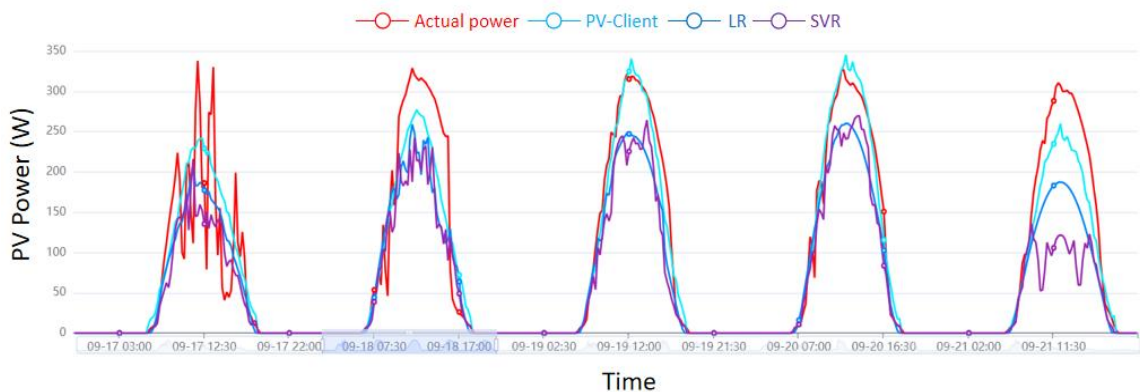

Fig. 4 PV power forecasting showcases of different models in Jingang Station.

3.3 Ablation study of PV-Client

In this section, ablation experiments are conducted to illustrate the effectiveness of our modifications to the Transformer module and the newly added modules for PV forecasting. The integrated linear module is expected to be particularly effective at capturing trend information in PV series, while the RevIN module plays a crucial role in addressing the issue of distribution shift in PV series. To evaluate the functions of the integrated linear module and RevIN, we examine the results after removing these modules from PV-Client, as presented in the '- Linear' and '- RevIN' columns in Table 4.

The embedding layer in the Transformer model usually serves to project the discrete input tokens into continuous vector representations, enabling the model to learn meaningful relationships between different tokens in the input. However, for time-series data, the input does not consist of discrete tokens. Instead, each time step exhibits strong correlations with other time steps. Therefore, introducing an additional embedding layer is speculated to compromise the inherent temporal information present in the PV series. To substantiate this claim, we assess the results of adding an embedding layer before the Encoder blocks, as shown in the '+ Embed' column in Table 4.

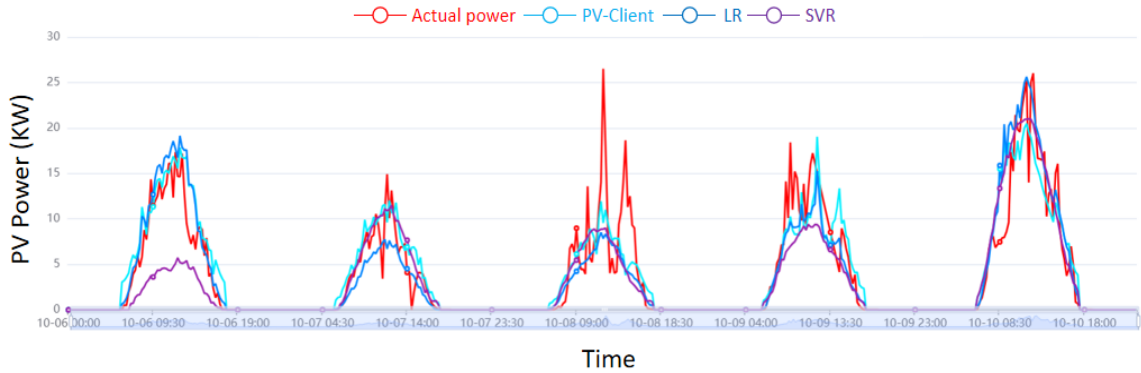

Fig. 5 PV power forecasting showcases of different models in Xinqingnian Station.

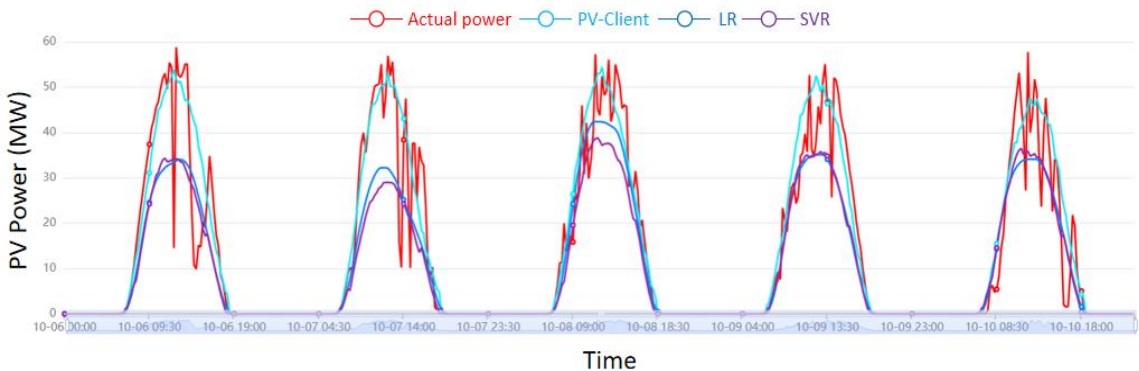

Fig. 6 PV power forecasting showcases of different models in Hongxing Station.

Furthermore, PV-Client replaces the Decoder component with a simple projection layer. We perceive the cross-variable based Transformer module in PV-Client to function more as a feature extractor than a series generator, considering the absence of a temporal relationship among different variables. Consequently, we consider a Decoder to be an unnecessary component in this context. To showcase that a projection layer is superior to a Decoder, we examine the results of replacing the projection layer with a Decoder, presented in the '+ Decoder' column in Table 4.

The experiments of Table 4 highlight that modifications to the PV-Client's model architecture result in performance degradation across all three PV power stations. Among them, after removing the linear module, the average MSE of the model increased by 16.87% across the three stations, indicating the significance of the linear module in capturing trends. After removing the RevIN module, the average MSE of the model increased by 10.26% across the three stations, reflecting the importance of the RevIN module in stabilizing the PV series. Furthermore, the inclusion of a redundant embedding layer may lead to information loss in time series data, contributing to performance degradation. Additionally, the inclusion of an unnecessary Decoder component could complicate the model's structure, potentially leading to overfitting. The ablation experiments on PV-Client have demonstrated that both the linear and RevIN modules are indispensable for PV power forecasting, and it has been shown that removing the embedding layer and replacing the Decoder with a projection layer are also effective modifications.

Table 4. Ablation experiments of PV-Client in three stations. MSE metrics is adopted. The best result is in bold font.

|  | PV-Client | - Linear | - RevIN | + Embed | + Decoder |
|---|---|---|---|---|---|
| Jingang Station | **1469.132** | 1665.021 | 1508.758 | 1652.039 | 1844.109 |
| Xinqingnian Station | **6.396** | 10.121 | 8.787 | 9.753 | 11.214 |
| Hongxing Station | **72.741** | 74.234 | 73.497 | 75.393 | 75.172 |

*3.4 Additional Analysis of Client*

In this section, a more in-depth analysis of PV-Client is conducted. Firstly, the predictions of the integrated linear module and the Enhanced Transformer module in PV-Client are visualized to demonstrate that trends are captured by the linear module, while details are supplemented by the Transformer module. Next, an exploration is undertaken to determine whether the Attention mechanism is the optimal tool for learning the dependencies between PV power and weather factors. Furthermore, the impact of different historical lengths on the predictions of PV-Client is investigated. Finally, the impact of different options for the variable dimension of PV-Client's output is analyzed.

**Visualization of the prediction decomposition.** The integrated linear module is utilized to learn trends in the PV series, and the cross-variable Transformer module is employed to capture non-linear information and dependencies between PV power and weather factors, thereby complementing the details in the model's prediction. To substantiate this perspective, the PV-Client's prediction decomposition of the integrated linear module and the cross-variable Transformer module is presented separately, as shown in Fig. 7. It is observable that the predictions of the linear module effectively capture the trend in the PV series, while the predictions of the cross-variable Transformer module complement the details in the PV series.

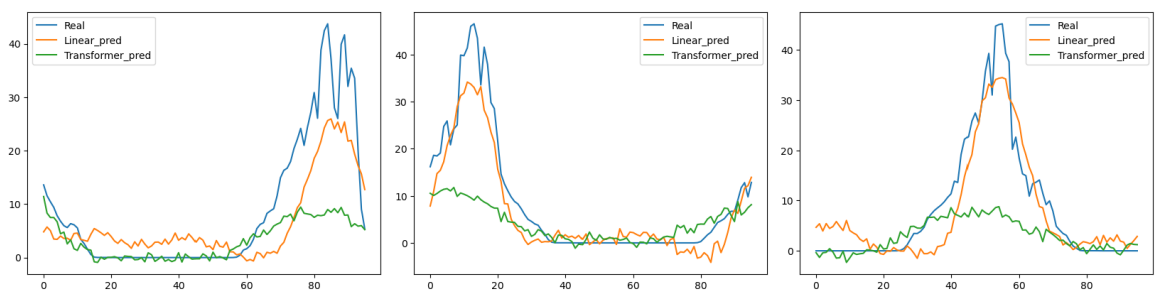

Fig. 7 Decomposition of PV-Client's predictions.

The overall concept of PV-Client is akin to that of TgDLF [32] and Adaptive-TgDLF [33]. All three methodologies involve the decomposition of energy (PV power or electricity) series into trend components and fluctuation (details) components. The primary distinction lies in the extraction of trend information. In the study of TgDLF and Adaptive-TgDLF, trend information is derived from domain knowledge, whereas in PV-Client, the trend information is autonomously learned by the integrated linear module.

**Comparison of the Attention module with alternative approaches.** The effectiveness of the Attention module in capturing dependencies between different variables has been demonstrated. We further test the capturing of dependencies between variables using alternative modules, such as ProbAttention employed in Informer [22], a linear layer, and an MLP layer. ProbAttention reduces the complexity of the original Attention by alternately sampling "key" and "query" and filling the missing values with the mean value. Additionally, we assess the performance of removing the Attention module from PV-Client, which is equivalent to solely enhancing the non-linear expressive capability of the linear model.

The experiments involving the replacement of the Attention module are conducted at Jingang Station, with the MSE metric being adopted. The results of these experiments are presented in Table 5. 'Attention' denotes the typical Attention module used in PV-Client. 'ProbAttention', 'Linear', and 'MLP' refer to replacing the typical Attention module with these alternative modules, while "No Attention" signifies the scenario where the Attention module is removed from PV-Client. The observations indicate that the model with the typical Attention module outperforms the ones with ProbAttention module, the linear layer, or MLP layer, as well as the model without the Attention module. Although ProbAttention improves computational efficiency, it also leads to information loss, resulting in a decline in model's prediction performance. This demonstration confirms the effectiveness of the typical Attention module in capturing dependencies between PV series and weather factors.

Table 5. The experiments of replacing the Attention module. Bold represents the best result.

| Dependencies-capturing module | MSE metrics |
|---|---|
| Attention | **1469.132** |
| ProbAttention | 1536.247 |
| Linear | 1769.433 |
| MLP | 1742.292 |
| No Attention | 1830.531 |

**Forecasting results with different historical lengths.** The length of the input historical PV series also affects the forecasting performance of the model. Generally, excessively short or long historical series are suboptimal. Excessively short PV power series may fail to capture sufficient trend information, leading to underfitting in deep-learning models. Conversely, excessively long PV power series may contain too much noise, potentially resulting in overfitting in deep-learning models. We test different lengths of historical PV input series at the three PV power stations, and the forecasting results are presented in Table 6. It is evident that a historical input duration of two days yields the best performance for PV-Client, which coincides with our basic experimental configuration. A one-day input may hinder the model's ability to learn adequately, while a four-day input risks overfitting, both resulting in inferior performance compared to the two-day input.

Table 6. PV-Client's forecasting results with different historical lengths. MSE metrics is adopted. The best result is indicated in bold font.

| | 96 (1 day) | 192(2 days) | 384(4 days) |
|---|---|---|---|
| Jingang Station | 1532.109 | **1469.132** | 1612.535 |
| Xinqingnian Station | 9.296 | **6.396** | 8.385 |
| Hongxing Station | 83.433 | **72.741** | 73.339 |

**Different options for the variable dimension of PV-Client's output.** Three alternatives exist for connecting the variable dimension of the PV-Client's output to the input dimensions. The first option entails a direct connection of the output variable dimension to the PV power dimension of the input, which is the method adopted in PV-Client. Alternatively, the second option involves associating the output variable dimension with the radiation forecast dimension of the input. Lastly, a third option involves computing a weighted sum of the outputs from both the PV power dimension and the radiation forecast dimension. For the third option, there are two possible implementations: one approach is to specify fixed weights for the summation of the two outputs (both set to 0.5 in our experiments); and the other approach treats the weights of the two outputs as learnable parameters, initially set at 0.5, and allows these weights to be updated along with the model during training.

We test these four different methods for determining the variable dimension of PV-Client's output, as shown in Table 7. 'PV dim / Radiation dim' refers to connecting the output variable dimension to either the PV power or radiation forecast dimension of the input. 'Sum (fixed weights) / Sum (updatable weights)' indicates computing a weighted sum of the outputs from both the PV power dimension and the radiation forecast dimension, with the weights either fixed or updatable. It is evident from the table that connecting the output variable dimension to the PV power dimension of the input yields the most effective results, which shows that PV-Client has learned accurate trend information from the historical PV power. Meanwhile, the other three options of determining the output variable dimension introduce potential biases from the radiation forecast data.

Table 7. PV-Client's forecasting results with variable dimensions of the output. MSE metrics is adopted. The best result is indicated in bold font.

|  | PV dim | Radiation dim | Sum (fixed weights) | Sum (updatable weights) |
| --- | --- | --- | --- | --- |
| Jingang Station | **1469.132** | 1639.660 | 1567.543 | 1506.173 |
| Xinqingnian Station | **6.396** | 10.445 | 8.983 | 7.996 |
| Hongxing Station | **72.741** | 82.917 | 76.438 | 73.247 |

## 4. CONCLUSION

In this study, we introduce PV-Client, an advanced and efficient model tailored for PV power forecasting. PV-Client leverages an Enhanced Transformer module for capturing cross-variable dependencies and a linear module for trend prediction in PV power series. Cross-variable Attention has proven to be effective in delineating associations between PV series and weather factors. This integrated approach empowers PV-Client to adeptly capture both global and local patterns in PV power, resulting in superior forecasting performance. The instance normalization and denormalization processes carried out by the RevIN module can further enhance the stability and forecasting capability of the PV-Client. Demonstrating SOTA outcomes on three real-world PV power stations, PV-Client exhibits notable advantages over both static methods and other deep learning methods. Specifically, PV-Client achieves a remarkable average of 6.3% improvement in MSE metrics, and an average of 0.9% enhancement in accuracy metrics across the three PV power stations. Importantly, PV-Client showcases versatility for seamless adaptation to various energy forecasting tasks, extending its applicability to domains such as wind power forecasting and electricity forecasting.

## CODE AVAILABILITY

The code for PV-Client is accessible for downloading through the following link: https://github.com/daxin007/PV-Client.

## DECLARATION OF COMPETING INTEREST

The authors assert that they have no competing financial interests or personal relationships that could have influenced the work reported in this paper.

## ACKNOWLEDGEMENTS

This work was supported by the National Natural Science Foundation of China (Grant No. 62106116), China Meteorological Administration under Grant QBZ202316, Natural Science Foundation of Ningbo of China (No. 2023J027), as well as by the High Performance Computing Centers at Eastern Institute of Technology, Ningbo, and Ningbo Institute of Digital Twin.